\documentclass[10pt,twocolumn,letterpaper]{article}
\usepackage{placeins}
\usepackage[pagenumbers]{cvpr}

\usepackage[dvipsnames]{xcolor}

\definecolor{cvprblue}{rgb}{0.21,0.49,0.74}
\usepackage[pagebackref,breaklinks,colorlinks,citecolor=cvprblue]{hyperref}

\def\paperID{*****}
\def\confName{CVPR}
\def\confYear{2024}

\title{Just Add Geometry: Gradient-Free Open-Vocabulary 3D Detection Without Human-in-the-Loop}

\usepackage{multirow}

\author{
  \begin{tabular}[t]{c@{\hskip 0.5in}c}
    Atharv Goel & Mehar Khurana \\
  \end{tabular}
  \\
  \texttt{\small \{atharv21027, mehar21541\}@iiitd.ac.in} \\\\
  Indraprastha Institute of Information Technology, Delhi
}

\begin{document}
\maketitle

\begin{abstract}

Modern 3D object detection datasets are constrained by narrow class taxonomies and costly manual annotations, limiting their ability to scale to open-world settings. In contrast, 2D vision-language models trained on web-scale image-text pairs exhibit rich semantic understanding and support open-vocabulary detection via natural language prompts. In this work, we leverage the maturity and category diversity of 2D foundation models to perform open-vocabulary 3D object detection without any human-annotated 3D labels.

Our pipeline uses a 2D vision-language detector to generate text-conditioned proposals, which are segmented with SAM and back-projected into 3D using camera geometry and either LiDAR or monocular pseudo-depth. We introduce a geometric inflation strategy based on DBSCAN clustering and Rotating Calipers to infer 3D bounding boxes without training. To simulate adverse real-world conditions, we construct \textbf{Pseudo-nuScenes}, a fog-augmented, RGB-only variant of the nuScenes dataset.

Experiments demonstrate that our method achieves competitive localization performance across multiple settings, including LiDAR-based and purely RGB-D inputs, all while remaining training-free and open-vocabulary. Our results highlight the untapped potential of 2D foundation models for scalable 3D perception. We open-source our code and resources at \href{https://github.com/atharv0goel/open-world-3D-det}{https://github.com/atharv0goel/open-world-3D-det}.

\end{abstract}    
\section{Introduction}
\label{sec:intro}

Object detection is a foundational task in computer vision, critical for applications such as autonomous driving \cite{caesar2020nuscenes}, augmented reality, and robotics~\cite{xiang2018posecnn}. While traditional approaches focus on detecting objects in 2D images~\cite{ren2016faster, redmon2016yolo}, the need for richer spatial understanding has led to the development of 3D object detection methods that operate on modalities like point clouds or RGB-D data~\cite{qi2018frustum, lang2019pointpillars}. These methods reason about object geometry and spatial layout in three-dimensional space, offering more precise localization and scene comprehension.

Despite this progress, 3D object detection remains constrained by the limited class taxonomies available in existing 3D datasets, which typically contain only a handful of object categories. For instance, KITTI~\cite{geiger2012kitti} includes just three classes, while nuScenes~\cite{caesar2020nuscenes} and Waymo~\cite{sun2020waymo} offer 10–23. In contrast, 2D detection datasets such as COCO~\cite{mscoco} or LVIS~\cite{gupta2019lvis} encompass hundreds or thousands of categories, providing a much broader object vocabulary. This discrepancy hampers the generalization capability of 3D detectors, particularly in open-world or real-world scenarios where object classes are diverse, fine-grained, and continually evolving~\cite{zareian2021open}.

Open-vocabulary object detection (OVD) addresses this limitation by enabling models to detect novel object categories not explicitly labeled during training. Instead of relying on fixed class labels, OVD methods typically leverage vision-language models trained on image-text pairs to associate regions with arbitrary textual queries~\cite{zareian2021open, OV_3DET, du2022glip}. Recent 2D OVD systems such as GLIP~\cite{du2022glip}, Grounding DINO~\cite{liu2023grounding}, and RegionCLIP~\cite{zhong2022regionclip} demonstrate impressive flexibility and scalability by learning rich semantic representations capable of grounding object categories from natural language.

In this work, we propose a method for open-vocabulary 3D object detection that eliminates the need for manual 3D annotations. Our approach builds upon large-scale open-vocabulary vision-language models trained on 2D images, using their 2D predictions as supervisory signals for 3D detection. This strategy is motivated by two key insights: (i) 2D datasets offer vastly greater object category coverage than their 3D counterparts~\cite{gupta2019lvis, mscoco}, and (ii) 2D open-vocabulary detectors are significantly more mature and scalable~\cite{du2022glip, liu2023grounding}.

\begin{figure*}[htb]
    \centering
    \includegraphics[width=1\linewidth]{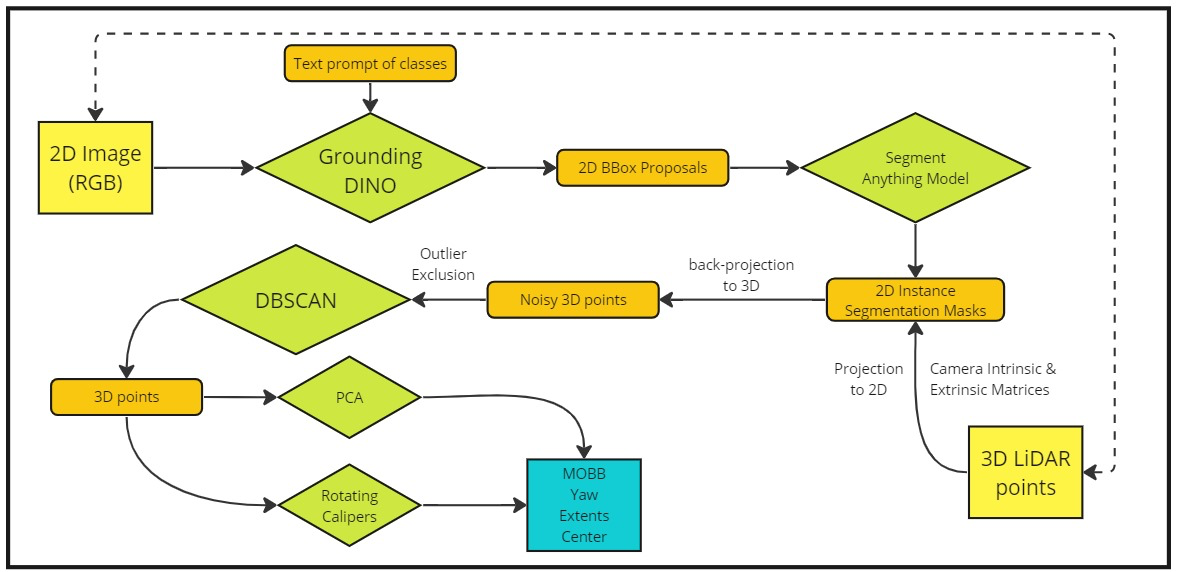}
    \caption{Pipeline of our methodology.}
    \label{fig:flowchart}
\end{figure*}

Without training any new model, our method transfers the generalization ability of 2D vision-language models to the 3D domain. This enables open-world 3D object detection in previously unseen environments, offering a scalable path toward 3D perception without the burden of dense 3D annotations.

\textbf{Our key contributions are as follows}:
\begin{itemize}
    \item We propose a novel framework for \textit{open-vocabulary 3D object detection} that requires no manual 3D annotations, leveraging off-the-shelf 2D vision-language models to generate 3D bounding boxes.
    \item We design a robust 2D-to-3D projection pipeline that incorporates SAM-based segmentation~\cite{SAM_SegmentAnythingModel}, self-supervised noise filtering via DBSCAN, and classical geometric methods to infer 3D bounding box parameters.
    \item We explore the feasibility of monocular RGB-only 3D detection by introducing a pseudo-LiDAR pipeline using zero-shot depth estimation~\cite{piccinelli2024unidepth}.
    \item We propose a novel benchmark titled \textbf{Pseudo-nuScenes}, to evaluate performance under adverse weather conditions and absence of LiDAR data.
    \item We conduct comprehensive ablations comparing inflation strategies and demonstrate that our method can produce competitive open-vocabulary 3D predictions under realistic constraints, without any training or dataset-specific priors.
\end{itemize}
\section{Related Work}
\label{sec:related-work}

\paragraph{Open-Vocabulary 2D Object Detection.}
Open-vocabulary detection (OVD) has gained traction as a way to break free from fixed taxonomies in object detection. Methods like GLIP~\cite{du2022glip}, Grounding DINO~\cite{liu2023grounding}, and RegionCLIP~\cite{zhong2022regionclip} use large-scale image-text pretraining to associate object regions with arbitrary text queries. These models have demonstrated remarkable zero-shot generalization, making them a powerful tool for object discovery. Our method builds on this foundation using 2D open-vocabulary detectors as the starting point for 3D localization, but differs by targeting 3D reasoning without training or fine-tuning any model.

\paragraph{3D Object Detection and Dataset Limitations.}
Conventional 3D object detection methods rely on annotated LiDAR or RGB-D point clouds~\cite{qi2018frustum, lang2019pointpillars, sun2020waymo}. These models are constrained by small taxonomies (e.g., 10–23 classes in nuScenes~\cite{caesar2020nuscenes}, Waymo~\cite{sun2020waymo}, or KITTI~\cite{geiger2012kitti}), and require costly annotation pipelines. Our approach avoids these limitations by leveraging 2D detections to supervise 3D localization, without requiring any 3D-labeled data.

\paragraph{Open-Vocabulary 3D Detection.}
A growing body of work explores bringing open-vocabulary capabilities to 3D.  
Lu \textit{et al.}~\cite{lu2023open} propose a training-based pipeline that uses 2D open-vocabulary models to generate pseudo-labels and then trains a 3D detector using cross-modal contrastive learning between point clouds, text, and images.  
CLIP2Scene~\cite{clip2scene} extends CLIP to 3D by transferring knowledge via 2D-to-3D alignment for indoor scenes.  
CLIP-FO3D~\cite{clipfo3d} proposes a frozen 3D backbone and aligns 3D proposals with CLIP embeddings for open-set 3D detection.  
FSD~\cite{wang2023fsd} incorporates CLIP into 3D detectors using a learned 3D-text similarity metric.

Compared to these methods, our pipeline is training-free, does not require any 3D proposals or feature extractors, and operates purely via geometric projection from 2D outputs. Most of the above methods require labeled training data in at least one modality or fine-tuning of encoders, whereas our system is entirely inference-time and model-agnostic.

\paragraph{Zero-Shot 3D Detection via Geometry.}
Wilson \textit{et al.}~\cite{wilson20203d} (3D for Free) use 2D segmentations and HD maps to generate 3D pseudo-annotations, which are then used to train a 3D detector. However, their method depends on dataset-specific priors and closed-set taxonomies, and is not designed for open-world settings. In contrast, our method uses open-vocabulary 2D outputs, requires no training or HD maps, and generalizes to novel object categories via natural language.

\paragraph{Vision-Language Models for 3D Understanding.}
Several works use CLIP-like models for 3D representation learning.  
PointCLIP~\cite{zhang2022pointclip} adapts CLIP to classify point clouds via projection.  
ULIP~\cite{xue2022ulip} aligns point cloud, image, and language embeddings into a unified space via contrastive learning.  
While related in spirit, these methods primarily focus on classification and require extensive multi-modal pretraining. In contrast, our approach treats pretrained 2D vision-language models as fixed oracles and focuses on fully automated 3D detection.

\paragraph{Summary.}
In summary, our method is the first to demonstrate fully training-free, open-vocabulary 3D object detection using only 2D foundation models and classical geometric reasoning. It avoids the need for dataset-specific priors, annotations, or learned 3D feature extractors, enabling scalable 3D perception from purely 2D supervision.

\section{Methodology}\label{sec:methodology}

We propose a zero-shot pipeline for open-vocabulary 3D object detection that leverages 2D vision-language models and 3D geometric reasoning, while requiring no human-labeled 3D annotations. Our approach is summarized in Figure~\ref{fig:flowchart} and consists of five key stages: 2D open-vocabulary detection, instance segmentation, 2D-to-3D back-projection, geometric box inflation, and optional pseudo-depth substitution.

\subsection{Stage 1: 2D Open-Vocabulary Detection}

Given a text prompt specifying target object classes, we first run a vision-language detector such as GroundingDINO~\cite{GroundingDINO} on the input RGB image. The model returns a set of text-conditioned 2D bounding boxes with associated class labels and confidence scores. These detections serve as the initial priors for subsequent 3D reasoning.

\subsection{Stage 2: Instance Segmentation via SAM}

To improve spatial localization and minimize noise in 3D point selection, we apply Segment Anything (SAM)~\cite{SAM_SegmentAnythingModel} to generate high-quality instance masks. We condition SAM on the bounding boxes predicted by GroundingDINO, producing fine-grained object-level segmentation masks.

\subsection{Stage 3: LiDAR-to-Image Projection and Back-Projection}

Next, we project the LiDAR point cloud into the image plane using the known camera intrinsics and extrinsics. This establishes a pixel-wise mapping between 2D pixels and 3D points. For each segmentation mask, we retrieve the corresponding 3D points by inverting this mapping, effectively \textit{lifting} the 2D instance mask into a 3D point cloud segment. However, this set of points may be noisy due to projection ambiguities and imperfections in the mask.

\subsection{Stage 4: 3D Bounding Box Inference}

To convert the masked 3D point cloud into a 3D bounding box, we explore several inflation strategies:

\begin{itemize}
    \item \textbf{Medoid-based Centering:} We compute the medoid of the 3D points as a robust estimate of the box center.
    \item \textbf{Rotating Calipers for Orientation and Size:} We adapt the classic Rotating Calipers algorithm to fit a minimum-area oriented bounding box to the point cloud in the ground plane. The box’s height is derived from the vertical extent of the points.
\end{itemize}

We compare the effectiveness of various combinations of these strategies (e.g., medoid center with shape priors vs. full Calipers inflation) in Section~\ref{sec:experiments}.

To reduce the impact of outliers, we apply DBSCAN clustering on the back-projected 3D points and retain only the densest cluster, yielding cleaner box proposals.

\subsection{Stage 5: Pseudo-Depth Variant (RGB-Only)}

In scenarios where LiDAR is unavailable, we substitute real depth data with pseudo-LiDAR generated from RGB images using UniDepth~\cite{piccinelli2024unidepth}, a zero-shot monocular depth estimator. We back-project the RGB pixels into 3D using the estimated depth map and intrinsic parameters. The rest of the pipeline remains identical. We evaluate this variant on our augmented Pseudo-NuScenes dataset, described in Section \ref{sec:psuedo-nuscenes}.

\subsection{Label Assignment}

For each predicted 3D box, we transfer the class label and confidence score from the original 2D detection. No additional training or fine-tuning is performed, and our pipeline remains entirely model-agnostic with respect to the downstream 3D task.

\begin{table*}[htbp]
    \centering
    \caption{Comparison of inflation strategies on \textbf{nuScenes}. Metrics are reported on the 5 most common classes.}\label{nusc-table}
    \label{tab:inflation-table}
    \begin{tabular}{lccccccccc}
        \toprule
        \multirow{1}{*}{Method} & DBSCAN & mAP & mATE & mASE & mAOE & mAVE & mAAE & mAR & NDS\\

        \midrule
        {Medoid + Lane geometry} 
        & Yes & \textbf{29.94\%} & 0.938 & 0.700 & \textbf{1.045} & 1.560 & 0.982 & 41.78\% & \textbf{18.77\%}\\
        { + shape priors}
         & No & 29.42\% & 0.948 & 0.700 & 1.045 & 1.558 & 0.982 & 40.24\% & 18.41\%\\
        \midrule
        {Rot. Calipers (center,} 
        & Yes & \textbf{21.94\%} & 0.956 & 0.879 & \textbf{1.155} & 1.566 & 0.980 
        & 36.74\% & \textbf{12.82\%}\\
        { orientation, shape) }
         & No & 1.30\% & 1.029 & 0.977 & 1.144 & 1.151 & 0.990 & 6.76\% & 0.99\% \\
         \midrule
        {Medoid + Rotating Calipers for} 
        & Yes & \textbf{29.30\%} & 0.949 & 0.897 & \textbf{1.155} & 1.552 & 0.981 
        & 40.10\% & \textbf{16.38\%}\\
        {orientation, shape}
         \\
         \midrule
        \multirow{1}{*}{3D For Free w/ HD maps} 
        & - & 37.40\% & 0.41 & 0.31 & 0.90 \\
        \midrule
        \multirow{1}{*}{3D For Free w/ Rot. Calipers } 
        & - & 34.31\% & 0.54 & 0.33 & 1.35\\
        \bottomrule
    \end{tabular}
\end{table*}

\section{Pseudo-nuScenes: Fog-Augmented RGB-D Benchmark}\label{sec:psuedo-nuscenes}

Recent research in 3D object detection has highlighted the limitations of relying solely on LiDAR data: high cost, limited accessibility, and sensitivity to weather. Moreover, most existing datasets fail to account for real-world degradations such as fog or occlusion. To study the effectiveness of open-vocabulary 3D object detection in such conditions, we introduce a novel benchmark: \textbf{Pseudo-nuScenes}.

\subsection{Dataset Construction}

Pseudo-nuScenes is derived from the official mini split of the nuScenes dataset~\cite{nuscenes}, which contains over 1000 annotated frames from synchronized RGB and LiDAR sensors. We discard all LiDAR signals and instead generate a pseudo-3D structure from monocular RGB using state-of-the-art zero-shot metric depth estimation.

\textbf{Monocular Depth via UniDepth.} For each RGB image, we use UniDepth~\cite{piccinelli2024unidepth}, a universal monocular depth model, to infer metric depth maps. These depth maps are projected into 3D using known intrinsic parameters from the nuScenes calibration data. The resulting pseudo-point clouds mimic LiDAR scans and can be processed by our 3D inflation pipeline as drop-in replacements.

\textbf{Depth Generation.} For each RGB image from the six nuScenes cameras, we run UniDepth to obtain per-pixel metric depth maps. These maps are projected into 3D using the known camera intrinsics, producing dense pseudo-point clouds. The resulting 3D structure allows us to simulate LiDAR-like observations for each frame without requiring any additional sensor data.

\textbf{Fog Augmentation.} To simulate real-world visibility degradation, we apply a physics-inspired fog model to the RGB images. The model attenuates pixel intensities based on scene depth using the standard exponential transmittance function:
\[
I_{\text{fog}}(x) = I(x) \cdot t(x) + A \cdot (1 - t(x)), \quad t(x) = e^{-\beta d(x)}
\]
where \( I(x) \) is the original pixel intensity, \( d(x) \) is the depth at pixel \( x \), \( \beta \) is the fog density, and \( A \) is the ambient atmospheric light (set to white). This simulates realistic fog effects where distant regions become low contrast and desaturated.

\textbf{Setup.} We process all frames from the mini split of nuScenes, resulting in a fully RGB-D dataset that can be used to test open-vocabulary 3D detection pipelines under more challenging conditions. Our dataset is particularly useful for evaluating methods in domains where active sensors (e.g., LiDAR) are unavailable, unreliable, or prohibitively expensive.

\section{Experiments}
\label{sec:experiments}

\begin{table*}[htb]
\centering
\caption{Performance on Pseudo-nuScenes (fog + UniDepth).}
\label{tab:pseudo-nuscenes}
\begin{tabular}{lccccccc}
\toprule
\textbf{Method} & \textbf{mAP} & \textbf{mATE} & \textbf{mASE} & \textbf{mAOE} & \textbf{mAVE} & \textbf{mAAE} & \textbf{NDS} \\
\midrule
Medoid + Shape Priors + Lane Geometry & 16.14 & 1.053 & 0.703 & 1.039 & 1.545 & 0.981 & 11.23 \\
Rot. Calipers (center, orientation, shape) & 12.21 & 1.060 & 0.890 & 1.148 & 1.484 & 0.980 & 7.40 \\
\bottomrule
\end{tabular}
\end{table*}

\subsection{Datasets}

\paragraph{nuScenes.} We evaluate our method on the mini-split of the nuScenes dataset~\cite{nuscenes}, a large-scale multimodal benchmark for autonomous driving. The dataset includes synchronized RGB images from six cameras, LiDAR scans, and 3D bounding box annotations across 23 object classes. For our experiments, we use only the two validation sequences (10 scenes total) due to our model-free setup and compute constraints.

\paragraph{Pseudo-nuScenes (Ours).} We introduce a fog-augmented, RGB-D variant of nuScenes, named \textit{Pseudo-nuScenes}, to study the robustness of our method in RGB-only settings. We discard all real LiDAR and generate pseudo-LiDAR from RGB images using UniDepth~\cite{piccinelli2024unidepth}, a zero-shot monocular metric depth estimator. We additionally apply synthetic fog augmentations using a depth-aware haze simulation to mimic adverse weather. Our dataset allows us to evaluate open-vocabulary detection under degraded, real-world visual conditions.

\subsection{Evaluation Metrics}

We follow the nuScenes detection benchmark protocol~\cite{caesar2020nuscenes}. Our primary metric is:
\begin{itemize}
    \item \textbf{Mean Average Precision (mAP):} We follow the official definition, where a match is based on the 2D Euclidean distance of ground-plane box centers, averaged over thresholds $\{0.5, 1, 2, 4\}$ meters.
\end{itemize}

We also report the nuScenes suite of true positive (TP) quality metrics:
\begin{itemize}
    \item \textbf{mATE}: Mean Average Translation Error (m).
    \item \textbf{mASE}: Mean Average Scaling Error.
    \item \textbf{mAOE}: Mean Average Orientation Error (radians).
    \item \textbf{mAVE}: Mean Average Velocity Error (m/s).
    \item \textbf{mAAE}: Mean Average Attribute Error.
    \item \textbf{mAR}: Mean Average Recall.
    \item \textbf{NDS}: NuScenes Detection Score, combining all the above via a weighted average.
\end{itemize}

\subsection{Baselines and Variants}

We compare our method with:
\begin{itemize}
    \item \textbf{3D For Free}~\cite{wilson20203d}: A prior method that inflates 2D segmentations to 3D using HD maps and hand-crafted priors.
    \item \textbf{Ours (Medoid + Shape Priors)}: Uses LiDAR medoid as box center and fixed class-specific shape priors.
    \item \textbf{Ours (Rotating Calipers)}: Uses 3D Rotating Calipers for box orientation and size; we ablate center strategies.
    \item \textbf{Ours (Pseudo-Depth)}: Fully RGB-only pipeline using UniDepth + fog augmentation from Pseudo-nuScenes.
\end{itemize}

\section{Results and Discussion}\label{sec:results}

\subsection{Impact of Inflation Strategy}

Table~\ref{tab:inflation-table} presents results on the nuScenes mini validation set. We find that using the medoid as the center, along with handcrafted lane geometry and shape priors, achieves an mAP of 29.94\%, closely matching the performance of the prior 3D For Free baseline (34.31\%). Our method, in contrast, uses no dataset-specific priors or HD maps, and supports open-vocabulary queries.

We observe that the mASE scaling error increases by 0.2 between the shape priors baseline and both versions of rotating calipers. This increase is expected because the shape priors, by design, are hand-crafted features intended to generate more appropriately sized anchor boxes. Interestingly, our method achieves equivalent translation performance as the shape priors baseline. Recall, attribute error, and velocity error remain similar across the various methods as the inflation method has little consequence on these factors.

However, consider the orientation error (mAOE). Our method featuring the 3D Rotating Calipers strategy yields improved orientation estimates, with a lower mAOE (1.045 vs. 1.144), \textbf{surpassing the baseline} inflation methods \cite{wilson20203d}!

However, this method suffers in mAP and recall unless combined with medoid-based centering. This highlights the importance of combining robust shape estimation with reliable center inference.

\subsection{Noise Suppression via Clustering}

\begin{figure}[t]
    \centering
    \includegraphics[width=0.49\linewidth]{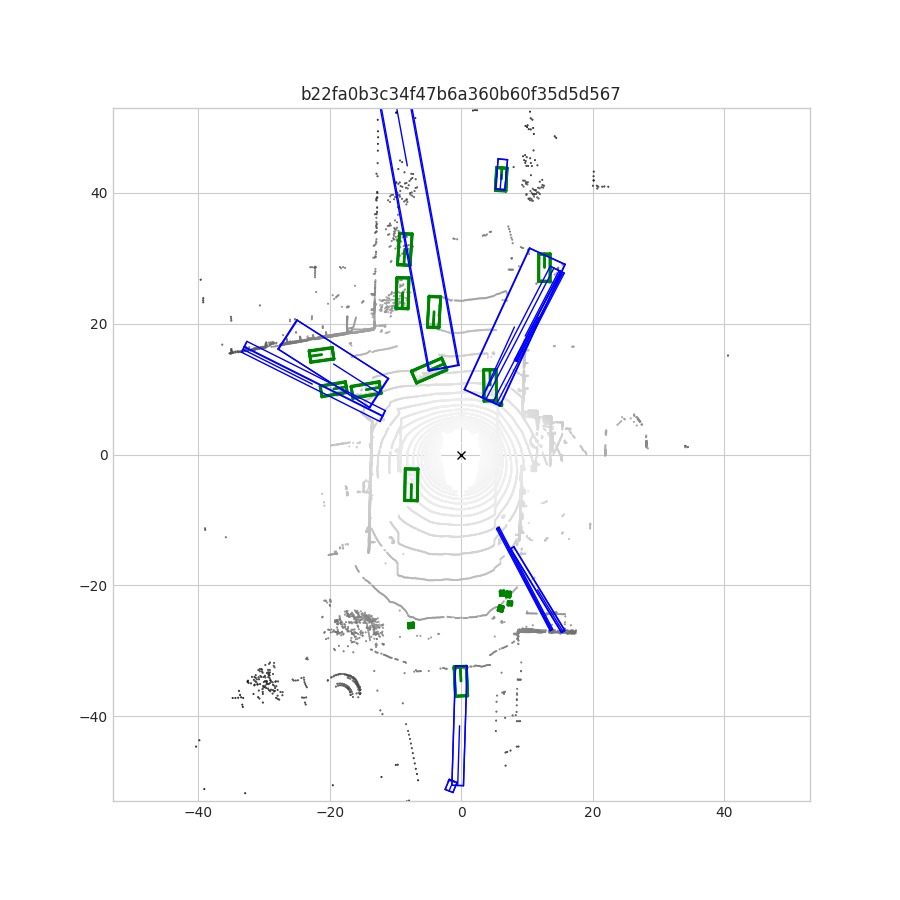}
    \includegraphics[width=0.49\linewidth]{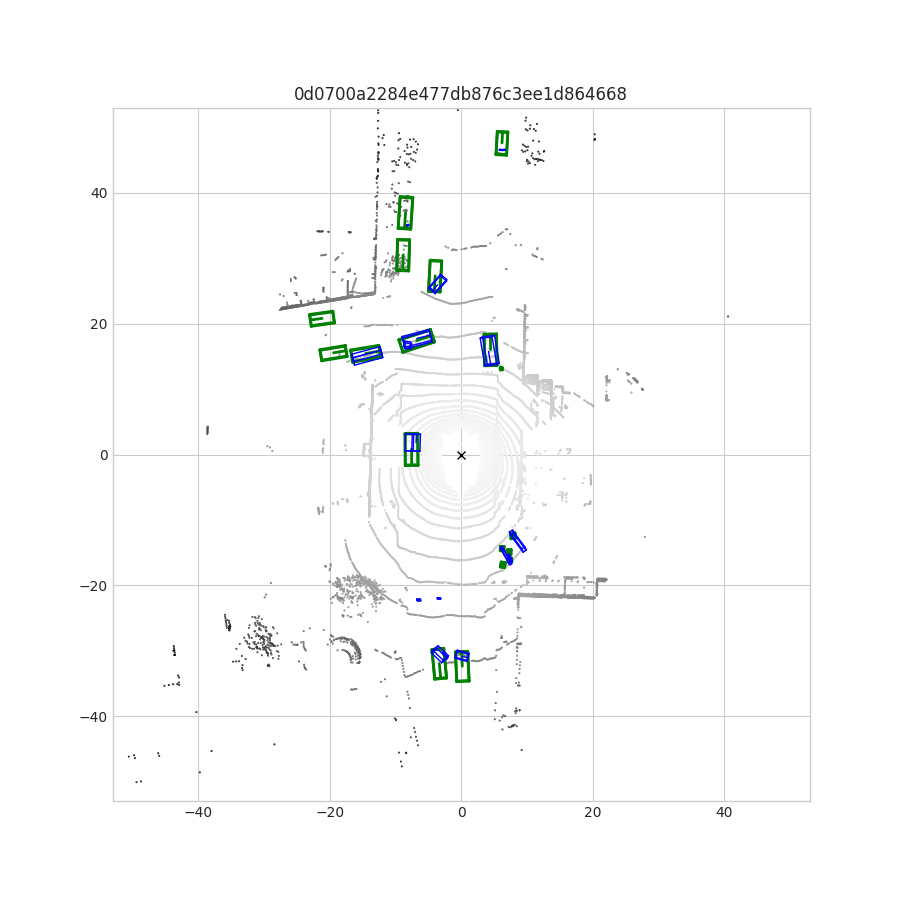}
    \caption{Effect of DBSCAN-based noise filtering on BEV maps. \textbf{Left}: Without filtering. \textbf{Right}: After DBSCAN. Blue: prediction, Green: ground truth.}
    \label{fig:noise-removal}
\end{figure}

To evaluate the effect of noise in projected 3D points, we apply DBSCAN for outlier removal. As shown in Figure~\ref{fig:noise-removal}, the noisy predictions result in severely degraded box quality. With DBSCAN, the mAP improves from 1.30\% to 21.94\%, affirming the value of density-based filtering.

\subsection{Performance in RGB-Only Settings}

In Table~\ref{tab:pseudo-nuscenes}, we show results on our Pseudo-nuScenes dataset. As expected, replacing LiDAR with pseudo-depth leads to a drop in accuracy, especially in mAR and orientation. Nevertheless, our method achieves non-trivial mAP values (12–16\%), suggesting potential for fully vision-only setups in constrained environments.

However, the translation error (mATE) remains roughly similar between the pseudo-depth predictions and the actual depth predictions. Likewise, scaling error (mASE) is equivalent, implying that the predicted boxes have equivalent performance in terms of box scale and positioning. However, there is significantly less recall using pseudo-depth. The predicted depth labels appear to lead to noisier/more confusing model predictions than the real depth labels, leading to lower recall.

\subsection{Qualitative Results}

\begin{figure}[htbp]
    \centering
    \includegraphics[width=0.49\linewidth]{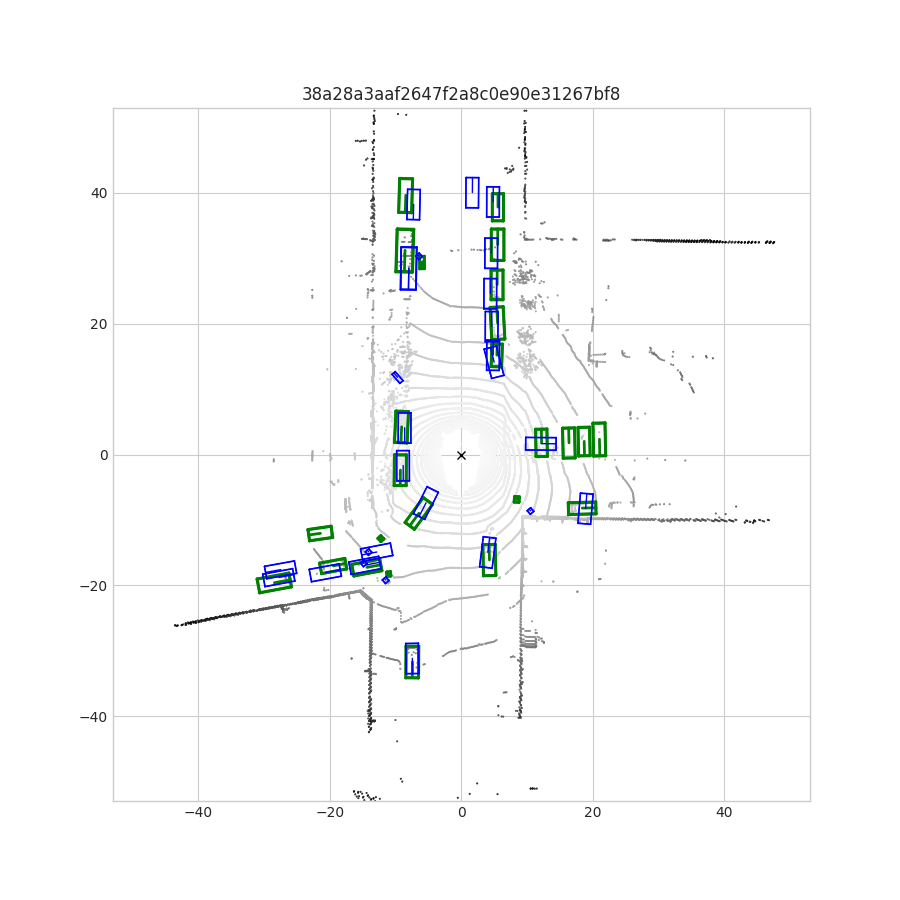}
    \includegraphics[width=0.49\linewidth]{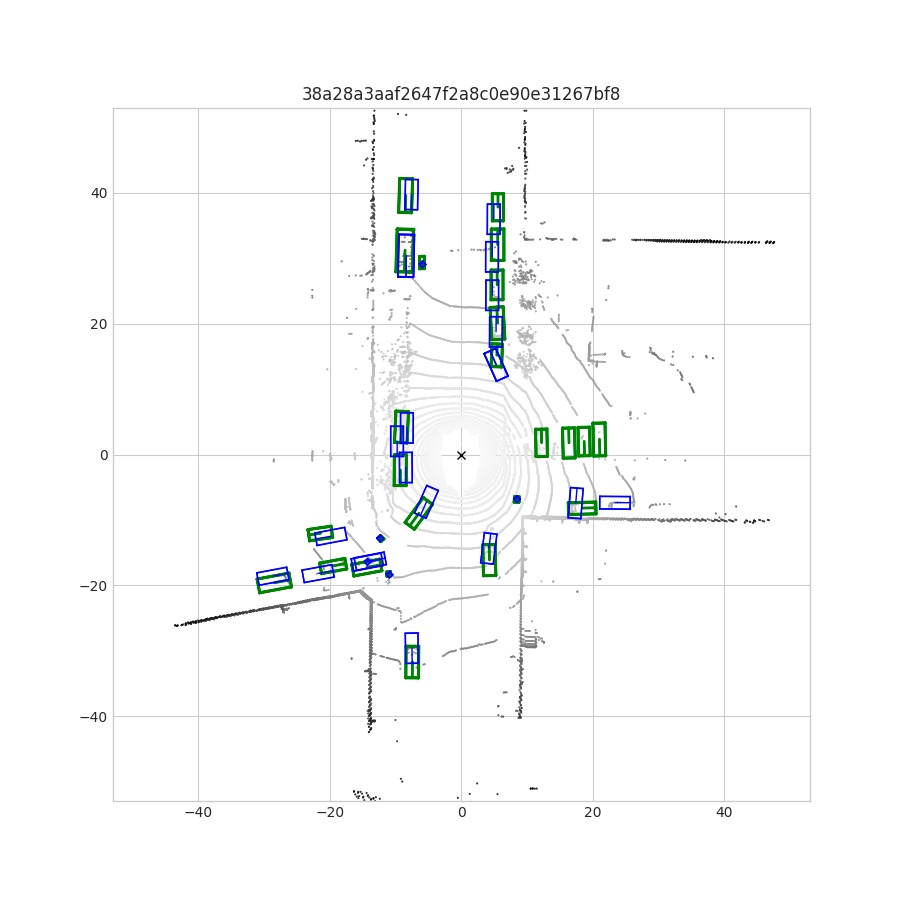}
    \caption{Qualitative results on nuScenes (left) and Pseudo-nuScenes (right).
    Blue: prediction, Green: ground truth.}
    \label{fig:qualitative}
\end{figure}

Figure~\ref{fig:qualitative} shows BEV visualizations. Our method captures major structures accurately in LiDAR-rich setups. In RGB-only settings, the predictions become noisier but still capture coarse layout.

\section{Conclusion}\label{sec:conclusion}

We present a new paradigm for open-vocabulary 3D object detection that mitigates the need for human-labeled 3D annotations. The core principle underlying our work is that 2D vision-language models trained on web-scale, richly diverse datasets possess semantic understanding and object localization capabilities that far exceed what current 3D datasets and detectors can offer. By harnessing these models as the supervisory backbone, we lift 2D detections to 3D space using classical geometric reasoning and minimal prior assumptions. Additionally, using our method leverages the larger taxonomies and maturity of 2D datasets.

Our pipeline remains entirely training-free, modular, and open-vocabulary, supporting flexible object queries and generalization to novel categories. To evaluate the real-world viability of this approach, we also introduce \textbf{Pseudo-nuScenes}, a fog-augmented, RGB-only benchmark derived from nuScenes, which simulates degraded visibility and absence of depth sensors.

Extensive experiments show that:
\begin{itemize}
    \item 2D detectors can serve as surprisingly strong priors for 3D object discovery,
    \item Our inflation strategy using segmentation masks, DBSCAN filtering, and Rotating Calipers achieves competitive localization performance,
    \item Monocular pseudo-depth provides a feasible (albeit noisy) alternative to LiDAR in open-world 3D settings.
\end{itemize}

By transferring the strengths of 2D detection to the 3D domain, our work takes a step toward scalable, annotation-free, open-vocabulary 3D perception. We hope this direction inspires future research on bridging the modality gap through 2D-centric supervision, especially as 2D foundation models continue to improve.

\section*{Acknowledgements}\label{sec:acknowledgement}

We would like to express our sincere gratitude to the Vision Lab at the Infosys Centre for AI, Indraprastha Institute of Information Technology, Delhi, for their generous support in providing us an NVIDIA A100 GPU for conducting our experiments. We also extend our thanks to Dr. Saket Anand for his mentorship as our bachelor's thesis advisor. His support was instrumental in helping us develop the skills necessary to execute this independent research.
{
    \small
    \bibliographystyle{ieeenat_fullname}
    \bibliography{main}

\begin{thebibliography}{26}
\providecommand{\natexlab}[1]{#1}
\providecommand{\url}[1]{\texttt{#1}}
\expandafter\ifx\csname urlstyle\endcsname\relax
  \providecommand{\doi}[1]{doi: #1}\else
  \providecommand{\doi}{doi: \begingroup \urlstyle{rm}\Url}\fi

\bibitem[Caesar et~al.(2020{\natexlab{a}})Caesar, Bankiti, Lang, Vora, Liong, Xu, Krishnan, Pan, Baldan, and Beijbom]{nuscenes}
Holger Caesar, Varun Bankiti, Alex~H. Lang, Sourabh Vora, Venice~Erin Liong, Qiang Xu, Anush Krishnan, Yu Pan, Giancarlo Baldan, and Oscar Beijbom.
\newblock nuscenes: A multimodal dataset for autonomous driving.
\newblock In \emph{CVPR}, 2020{\natexlab{a}}.

\bibitem[Caesar et~al.(2020{\natexlab{b}})Caesar, Bankiti, Lang, Vora, et~al.]{caesar2020nuscenes}
Holger Caesar, Varun Bankiti, Alex~H Lang, Sourabh Vora, et~al.
\newblock nuscenes: A multimodal dataset for autonomous driving.
\newblock In \emph{CVPR}, 2020{\natexlab{b}}.

\bibitem[Du et~al.(2022)Du, Zhang, Li, Lin, et~al.]{du2022glip}
Xiao Du, Hong Zhang, Zhen Li, Xiangyang Lin, et~al.
\newblock Glip: Grounded language-image pre-training.
\newblock In \emph{CVPR}, 2022.

\bibitem[Geiger et~al.(2012)Geiger, Lenz, and Urtasun]{geiger2012kitti}
Andreas Geiger, Philip Lenz, and Raquel Urtasun.
\newblock Are we ready for autonomous driving? the kitti vision benchmark suite.
\newblock In \emph{CVPR}, 2012.

\bibitem[Gupta et~al.(2019)Gupta, Dollar, and Girshick]{gupta2019lvis}
Agrim Gupta, Piotr Dollar, and Ross Girshick.
\newblock Lvis: A dataset for large vocabulary instance segmentation.
\newblock In \emph{Proceedings of the IEEE Conference on Computer Vision and Pattern Recognition}, 2019.

\bibitem[Kirillov et~al.(2023)Kirillov, Mintun, Ravi, Mao, Rolland, Gustafson, Xiao, Whitehead, Berg, Lo, Doll{\'a}r, and Girshick]{SAM_SegmentAnythingModel}
Alexander Kirillov, Eric Mintun, Nikhila Ravi, Hanzi Mao, Chloe Rolland, Laura Gustafson, Tete Xiao, Spencer Whitehead, Alexander~C. Berg, Wan-Yen Lo, Piotr Doll{\'a}r, and Ross Girshick.
\newblock Segment anything.
\newblock \emph{arXiv:2304.02643}, 2023.

\bibitem[Lang et~al.(2019)Lang, Vora, Caesar, Zhou, Yang, and Beijbom]{lang2019pointpillars}
Alex~H Lang, Sourabh Vora, Holger Caesar, Lubing Zhou, Jiong Yang, and Oscar Beijbom.
\newblock Pointpillars: Fast encoders for object detection from point clouds.
\newblock In \emph{CVPR}, 2019.

\bibitem[Li et~al.(2023)Li, Ye, Wang, et~al.]{clipfo3d}
Zhengxiong Li, Qian Ye, Tianrui Wang, et~al.
\newblock Clip-fo3d: Free open-vocabulary 3d object detection.
\newblock In \emph{ICCV}, 2023.

\bibitem[Lin et~al.(2014)Lin, Maire, Belongie, Hays, Perona, Ramanan, Doll{\'a}r, and Zitnick]{mscoco}
Tsung-Yi Lin, Michael Maire, Serge Belongie, James Hays, Pietro Perona, Deva Ramanan, Piotr Doll{\'a}r, and C.~Lawrence Zitnick.
\newblock Microsoft coco: Common objects in context.
\newblock In \emph{Computer Vision -- ECCV 2014}, pages 740--755, Cham, 2014. Springer International Publishing.

\bibitem[Liu et~al.(2023{\natexlab{a}})Liu, Zeng, Ren, Li, Zhang, Yang, Li, Yang, Su, Zhu, et~al.]{GroundingDINO}
Shilong Liu, Zhaoyang Zeng, Tianhe Ren, Feng Li, Hao Zhang, Jie Yang, Chunyuan Li, Jianwei Yang, Hang Su, Jun Zhu, et~al.
\newblock Grounding dino: Marrying dino with grounded pre-training for open-set object detection.
\newblock \emph{arXiv preprint arXiv:2303.05499}, 2023{\natexlab{a}}.

\bibitem[Liu et~al.(2023{\natexlab{b}})Liu, Zeng, Ren, Li, Zhang, et~al.]{liu2023grounding}
Shilong Liu, Zhaoyang Zeng, Tianhe Ren, Feng Li, Hao Zhang, et~al.
\newblock Grounding dino: Marrying dino with grounded pre-training for open-set object detection.
\newblock \emph{arXiv preprint arXiv:2303.05499}, 2023{\natexlab{b}}.

\bibitem[Liu et~al.(2023{\natexlab{c}})Liu, Xu, et~al.]{clip2scene}
Yujing Liu, Wenhao Xu, et~al.
\newblock Clip2scene: Scene-level 3d open-world understanding via vision-language foundation models.
\newblock In \emph{CVPR}, 2023{\natexlab{c}}.

\bibitem[Lu et~al.(2023{\natexlab{a}})Lu, Xu, Wei, Xie, Tomizuka, Keutzer, and Zhang]{OV_3DET}
Yuheng Lu, Chenfeng Xu, Xiaobao Wei, Xiaodong Xie, Masayoshi Tomizuka, Kurt Keutzer, and Shanghang Zhang.
\newblock Open-vocabulary point-cloud object detection without 3d annotation.
\newblock \emph{Proceedings of the IEEE/CVF Conference on Computer Vision and Pattern Recognition}, 2023{\natexlab{a}}.

\bibitem[Lu et~al.(2023{\natexlab{b}})Lu, Xu, Wei, Xie, Tomizuka, Keutzer, and Zhang]{lu2023open}
Yuheng Lu, Chenfeng Xu, Xiaobao Wei, Xiaodong Xie, Masayoshi Tomizuka, Kurt Keutzer, and Shanghang Zhang.
\newblock Open-vocabulary point-cloud object detection without 3d annotation.
\newblock In \emph{Proceedings of the IEEE/CVF Conference on Computer Vision and Pattern Recognition}, pages 1190--1199, 2023{\natexlab{b}}.

\bibitem[Piccinelli et~al.(2024)Piccinelli, Yang, Sakaridis, Segu, Li, Van~Gool, and Yu]{piccinelli2024unidepth}
Luigi Piccinelli, Yung-Hsu Yang, Christos Sakaridis, Mattia Segu, Siyuan Li, Luc Van~Gool, and Fisher Yu.
\newblock Unidepth: Universal monocular metric depth estimation.
\newblock In \emph{IEEE Conference on Computer Vision and Pattern Recognition (CVPR)}, 2024.

\bibitem[Qi et~al.(2018)Qi, Liu, Wu, Su, and Guibas]{qi2018frustum}
Charles~R Qi, Wei Liu, Chen Wu, Hao Su, and Leonidas~J Guibas.
\newblock Frustum pointnets for 3d object detection from rgb-d data.
\newblock In \emph{CVPR}, 2018.

\bibitem[Redmon et~al.(2016)Redmon, Divvala, Girshick, and Farhadi]{redmon2016yolo}
Joseph Redmon, Santosh Divvala, Ross Girshick, and Ali Farhadi.
\newblock You only look once: Unified, real-time object detection.
\newblock In \emph{Proceedings of the IEEE conference on computer vision and pattern recognition}, pages 779--788, 2016.

\bibitem[Ren et~al.(2016)Ren, He, Girshick, and Sun]{ren2016faster}
Shaoqing Ren, Kaiming He, Ross Girshick, and Jian Sun.
\newblock Faster r-cnn: Towards real-time object detection with region proposal networks.
\newblock \emph{IEEE transactions on pattern analysis and machine intelligence}, 39\penalty0 (6):\penalty0 1137--1149, 2016.

\bibitem[Sun et~al.(2020)Sun, Kretzschmar, et~al.]{sun2020waymo}
Pei Sun, Henrik Kretzschmar, et~al.
\newblock Scalability in perception for autonomous driving: Waymo open dataset.
\newblock In \emph{CVPR}, 2020.

\bibitem[Wang et~al.(2023)Wang, Zhou, Yang, et~al.]{wang2023fsd}
Yuchen Wang, Qi Zhou, Jianwei Yang, et~al.
\newblock Fsd: Few-shot object detection in 3d scenes.
\newblock In \emph{CVPR}, 2023.

\bibitem[Wilson et~al.(2020)Wilson, Kira, and Hays]{wilson20203d}
Benjamin Wilson, Zsolt Kira, and James Hays.
\newblock 3d for free: Crossmodal transfer learning using hd maps.
\newblock \emph{arXiv preprint arXiv:2008.10592}, 2020.

\bibitem[Xiang et~al.(2018)Xiang, Schmidt, Narayanan, and Fox]{xiang2018posecnn}
Yu Xiang, Tanner Schmidt, Venkatraman Narayanan, and Dieter Fox.
\newblock Posecnn: A convolutional neural network for 6d object pose estimation in cluttered scenes.
\newblock In \emph{PoseCNN: A Convolutional Neural Network for 6D Object Pose Estimation in Cluttered Scenes}, 2018.

\bibitem[Xue et~al.(2022)Xue, Wang, Liu, et~al.]{xue2022ulip}
Yujing Xue, Yue Wang, Xingyu Liu, et~al.
\newblock Ulip: Learning unified representation of language, image and point cloud for 3d understanding.
\newblock In \emph{ECCV}, 2022.

\bibitem[Zareian et~al.(2021)Zareian, Wang, Mottaghi, Farhadi, and Chang]{zareian2021open}
Alireza Zareian, Kevin~D Wang, Roozbeh Mottaghi, Ali Farhadi, and Shih-Fu Chang.
\newblock Open-vocabulary object detection using captions.
\newblock In \emph{CVPR}, 2021.

\bibitem[Zhang et~al.(2022)Zhang, Xie, Dai, and Yu]{zhang2022pointclip}
Yujing Zhang, Enze Xie, Jiwen Dai, and Zhaoxiang Yu.
\newblock Pointclip: Point cloud understanding by clip.
\newblock In \emph{CVPR}, 2022.

\bibitem[Zhong et~al.(2022)Zhong, Yang, Zhang, Li, Codella, Li, Zhou, Dai, Yuan, Li, et~al.]{zhong2022regionclip}
Yiwu Zhong, Jianwei Yang, Pengchuan Zhang, Chunyuan Li, Noel Codella, Liunian~Harold Li, Luowei Zhou, Xiyang Dai, Lu Yuan, Yin Li, et~al.
\newblock Regionclip: Region-based language-image pretraining.
\newblock In \emph{Proceedings of the IEEE/CVF Conference on Computer Vision and Pattern Recognition}, pages 16793--16803, 2022.

\end{thebibliography}
}

\end{document}